\documentclass[manuscript,screen]{acmart}
\AtBeginDocument{%
  \providecommand\BibTeX{{%
    \normalfont B\kern-0.5em{\scshape i\kern-0.25em b}\kern-0.8em\TeX}}}
    
\usepackage{tikz}
\usepackage{subcaption}
\usepackage{natbib}
\usepackage{listings}
\usepackage{mathtools}

\usepackage{amsthm}

\copyrightyear{2023}
\acmYear{2023}
\setcopyright{acmlicensed}\acmConference[FDG 2023]{Foundations of Digital Games 2023}{April 11--14, 2023}{Lisbon, Portugal}
\acmBooktitle{Foundations of Digital Games 2023 (FDG 2023), April 11--14, 2023, Lisbon, Portugal}
\acmPrice{15.00}
\acmDOI{10.1145/3582437.3582484}
\acmISBN{978-1-4503-9855-8/23/04}

\usepackage{xcolor}

\newcommand*{\mathabxbfamily}{\fontencoding{U}\fontfamily{mathb}\selectfont}
\DeclareFontFamily{U}{mathb}{\hyphenchar\font45}
\DeclareFontShape{U}{mathb}{m}{n}{
      <5> <6> <7> <8> <9> <10> gen * mathb
      <10.95> mathb10 <12> <14.4> <17.28> <20.74> <24.88> mathb12
      }{}

\newcommand*{\Neptune}{{\text{\mathabxbfamily\char"48}}}

\newcommand*{\wasyfamily}{\fontencoding{U}\fontfamily{wasy}\selectfont}

\newcommand*{\mercury}{{\text{\wasyfamily\char39}}}

\makeatletter
\def\@fnsymbol#1{\ensuremath{
\ifcase#1\or
\mercury \or  
\Neptune \or 
\mathsection \or
\mathparagraph \or
\|\or **
\or
\dagger\dagger
\or \ddagger\ddagger
\else\@ctrerr\fi}}
\makeatother

\begin{document}

\title{Why Oatmeal is Cheap: Kolmogorov Complexity and Procedural Generation}


\author{Younès Rabii}
\affiliation{
  \institution{Queen Mary of London}
  \country{United Kingdom}}
\email{yrabii.eggs@gmail.com}

\authornote{References to this author must be made using the they/them singular neutral pronouns.}

\author{Michael Cook}
\affiliation{
  \institution{King's College London}
  \country{United Kingdom}}
\email{mike@gamesbyangelina.org}
\authornote{References to this author must be made using the he/him masculine or they/them singular neutral pronouns.}

\renewcommand{\shortauthors}{Rabii and Cook}

\begin{abstract}
Although procedural generation is popular among game developers, academic research on the topic has primarily focused on new applications, with some research into empirical analysis. In this paper we relate theoretical work in information theory to the generation of content for games. We prove that there is a relationship between the Kolomogorov complexity of the most complex artifact a generator can produce, and the size of that generator’s possibility space. In doing so, we identify the limiting relationship between the knowledge encoded in a generator, the density of its output space, and the intricacy of the artifacts it produces. We relate our result to the experience of expert procedural generator designers, and illustrate it with some examples.
  
\end{abstract}




\maketitle

\section{Introduction}
Procedural content generation is a well-established part of the modern game developer's toolkit. The Game Developer's Conference, the largest event in the games industry, has hosted over 50 talks in the last decade about procedural generation, from small-scale independent speakers to large AAA companies, covering disciplines from programming to art to writing. Correspondingly, procedural generation has been an increasingly hot topic among game AI researchers in the last two decades. The Procedural Generation Workshop at FDG, now in its twelfth year, is one of the longest-running workshops in the field of game AI, and dedicated paper tracks at conferences are a regular occurrence.

Despite the huge importance of content generation, and the wealth of time invested into developing practical techniques, the analysis of procedural generators is a relatively underdeveloped area of study. A few notable techniques have emerged over the last two decades of research \cite{smith10, summerville}, as well as studies of efficacy \cite{metrics, metrics2}, but many of the techniques used by game researchers have changed little in that time. As a result, a lot of procedural generation work is done by `feel', with postmortems shared at events such as the Roguelike Celebration\footnote{https://www.roguelike.club/} that indicate a successful approach that others can attempt to replicate or build upon. 

Attempts to abstract and generalise knowledge about game development are important, because they allow us to connect the dots between disparate games, designers and techniques. They can also help provide support and evidence, or even proof, of `folk myths' about AI, or phenomena that are reported by many game developers but have never been concretised. Understanding where these feelings come from, and relating them to established ideas from computer science, leads to new discoveries and a deeper understanding of the craft.

In this paper we relate \textit{Kolmogorov Complexity}, a concept from information theory, to the generation of content for games. We prove that there is a fixed relationship between the Kolomogorov complexity of the most complex artifact a generator can produce, and the size of that generator's possibility space. In doing so, we identify the limiting relationship between the knowledge encoded in a generator, the density of its output space, and the intricacy of the artifacts it produces. We relate our result to the folklore of procedural generators, and illustrate it with examples.

\section{Background}
\subsection{Kolmogorov Complexity}
Information theory is the study of how information is communicated and represented, the origins of which predate the development of programming languages. Information theory is often used to analyse programs and computation, particularly as part of complexity theory and computability analysis.

In \cite{kolmogorov} the author introduces the notion of \textit{algorithmic complexity}, later named Kolmogorov complexity. The Kolmogorov complexity of an object is defined as the length of the shortest program that will produce that object when executed. For example, a string of one thousand zeros can be written as follows:
\begin{center}
\begin{lstlisting}[language=Python]
for i in range(0, 1000):
	print(0)
\end{lstlisting}
\end{center}
However, a string containing the first one thousand prime numbers would require a more complicated program to compute it. In general, the less compressible an object is, the higher its Kolmogorov complexity, as it requires more specific code to describe each part of the object. 

\subsection{Theory of Generative Systems}
The analysis of generative systems is a subject of study within game AI and far beyond in mathematical spaces. Fields relating to modelling, forecasting and probability all have some relationship to predicting the behaviour of complex or nondeterministic systems. 

In game AI more specifically, researchers have developed techniques for understanding the behaviour of procedural generators, most often for the purpose of building tools that can analyse and visualise this information for designers and developers. A prominent early example of this is Expressive Range Analysis (ERA) by Smith and Whitehead, in which they use repeated sampling of a content generator, and then plot metric data for sampled content on a histogram \cite{smith10}. This provides a way to visually understand the behaviour of the generator, provided metrics are chosen with care.

Summerville builds on this work in \cite{summerville} and suggests ways this can be expanded to use richer visualisations, particularly for the field of PCGML which requires a different approach to assessment. Cook et al also expand on ERAs in \cite{danesh}, building an assistive design tool that performs randomised analyses to show meta-level exploration of the design space of the generator itself. Their tool, Danesh, also provides several intelligent methods for tuning and changing a generator that account for the uneven fitness landscape and nonlinear behaviour of parameters to generative systems.

The choice of metrics used in assessing content is also the subject of study, as while analytical techniques are often general, they rely on specific metrics to provide domain-specific context that enables a deeper understanding of the quality of a particular generator's output. This is a major weakness of content generation analysis, as writing useful metrics is a difficult skill that requires a deep understanding of the application domain. Analyses of these metrics show a mixed success in predicting quality, and even for genre-specific metrics their general-purpose usability is not clear \cite{metrics2}.

All of the approaches listed in this section are empirical in nature, and require experimental analysis of output from the generator. These provide useful, practical techniques for developers to apply to their systems. In this paper, we attempt to complement this body of work by providing a result that is grounded in the underlying theory of generative systems. Expanding this work is important in providing a deeper understanding of how all generative programs function.

\subsection{Folk Wisdom and PCG} \label{sec:oatmeal}
Perhaps in part \emph{because} of its status as a technique associated with experimentation and interdisciplinary work, procedural generation has given rise to a strong community of practitioners that span academic research, the arts, the games industry and more besides. Over the last ten years many unofficial and informal events and communities have sprung up dedicated to generative software, especially in games, such as PROCJAM\footnote{https://www.procjam.com/} and Everything Procedural\footnote{https://everythingprocedural.com/}, as well as events with a strong focus on PCG such as the Roguelike Celebration. This sharing of practices and experiences with an idiosyncratic technology has given rise to a kind of folk wisdom about procedural generation, that combines humour, learned experiences and internalised knowledge.

One well-known example of this is \textit{The Ten Thousands Bowls of Oatmeal Problem}, a term coined by Kate Compton and now one of the best-known idioms among procedural generation practitioners. In this analogy, Compton likens procedurally generated content to bowls of oatmeal, and uses this to highlight the meaninglessness of appeals to variety or unpredictability which often accompany sales pitches related to procedural generation. Every bowl of oatmeal is unique, Compton explains, but that does not make them interesting or valuable. Designers use this to understand that procedural generation alone does not guarantee variety or interest, and that systems must be carefully designed to use generative methods as an expressive tool, rather than a solution in and of itself. One can imagine a similar sentiment being expressed in a less engaging and memorable way, particularly in the context of academic research which is often criticised for being overly formal. The elegance of the metaphor is surely crucial in enabling this message to be remembered and shared so widely.

Procedural generation practitioners also engage deeply with the frustrations and difficulties of working with the technology. One widely-shared tweet by game developer Orteil explains that ``thanks to procedural generation, I can produce twice the content in double the time''\footnote{https://tinyurl.com/orteilpcg22}. This tongue-in-cheek statement tells us a lot about the procedural generation community: that there is a sense of self-awareness; that there is an understanding of the myths that people tell about the technology; and that, despite this, the tweet author still enjoys working with PCG. 

One of our goals in this paper is to connect formal theoretical ideas about generative systems and programming to the intuition and internalised knowledge of procedural generation practitioners. In doing so, we hope we can strengthen these ideas and help the procedural generation community build on top of them, as well as encouraging further academic research in the area.


\section{Proof: Complexity and Limits of Generative Systems}
In this section we present a proof which relates the Kolmogorov complexity of generated artefacts, the algorithm that generates them, and the possibility space defined by that algorithm. We first define the terms used in our proof, state our theorem in those terms, and then describe the proof itself.

\begin{definition}
A \textit{program} is a finite binary string, $p \in \{0,1\}^{n} \text{ for } n \in \mathbb{N}$. An \textit{input table} is a finite binary string, $i \in \{0,1\}^{n} \text{ for } n \in \mathbb{N}$. An \textit{artefact} is a finite binary string, $a \in \{0,1\}^{n} \text{ for } n \in \mathbb{N}$.
\end{definition}

\begin{definition}
$|p|$ denotes the length of the finite binary string $p$. $\#S$ denotes the number of elements contained by the finite set $S$.
\end{definition}

\begin{definition}
A \textit{generator}, $G$, is a deterministic program, modelled as a finite function that maps an input table $i$, to an artefact, $a$. The \textit{possibility space} of $G$ (i.e. the set of all artefacts output by $G$) is denoted by $\pi(G)$.
\begin{align*}
\pi(G) = \{G(i) : i \in \{0,1\}^{n}, n \in \mathbb{N}  \}
\end{align*}
\end{definition}

\begin{definition}\label{def:gen}
A program is \textit{terminating} if it terminates and returns an output in finite time. 

A generator $G$ is \textit{ideal} if it is terminating and it satisfies the following properties:
\begin{itemize}
    \item \textit{Fixed Input Size}: $G$ accepts as input only binary strings of a fixed length, denoted by $input_G$.
    \item \textit{Injectivity}: $G$ is an injective function (that is, every input of length $input_G$ is associated with exactly one output, and distinct inputs are associated with distinct outputs).
\end{itemize}
\end{definition}

Note that most procedural generators are terminating, especially those used in the production of game content (either online or offline). However, this is not the case for the other two properties of ideal generators, \textit{Fixed Input Size} and \textit{Injectivity}. Nevertheless, any non-ideal generator can be straightforwardly transformed into an ideal generator. We describe this transformation in section \ref{sec:transform} as an appendix.

From the latter two properties of an ideal generator in Definition \ref{def:gen}, we can observe that the size of a generator's possibility space is directly related to the fixed size of its inputs:
\begin{align}
\#\pi(G) = 2^{input_G} \label{p:pspace}
\end{align}

This follows from the fact that every input must be associated with a unique output (by injectivity) and that $G$ must accept \textit{every} binary string of length $input_G$ and no other strings.

\begin{definition}
\textit{Kolmogorov complexity}, $K$, is a function that takes as input an artefact, $a$, and as output provides the length of the shortest combined program and input, $p$ and $i$ respectively, such that $p(i) = a$.
\begin{align}
K(a) = \min_{p,i : p(i) = a}{|p| + |i|}\ \label{}
\end{align}

$K^*$ is a function which takes as input a generator, $G$, and returns the largest Kolmogorov complexity of any artefact in $\pi(G)$. That is:
\begin{align}
K^*(G) = Max \{K(a) : a \in \pi(G)\} \label{p:kstar}
\end{align}

\end{definition}

In the following theorem, we place a lower and upper bound on $K^*(G)$ given an ideal generator $G$. Later, we demonstrate why this result is relevant to modern procedural generation theory, and give examples of its application.

\begin{theorem}
An ideal generator $G$ always satisfies the following inequality:
\begin{align*}
|G| + log_2(\#\pi(G)) \geq K^*(G) \geq log_2(\#\pi(G))
\end{align*}
\end{theorem}

\begin{proof}
Pick an arbitrary generator $G$, such that $G$ is ideal. We prove each inequality separately.\\

\textbf{Upper bound}: $|G| + log_2(\#\pi(G)) \geq K^*(G)$.

By the definition of Kolmogorov complexity:
\begin{align}
    \exists i \in \{0,1\}^{input_G} \;.\; |G| + |i| \geq K^*(G) 
\end{align}

Since $G$ is an ideal generator, $|i|$ is the same for all inputs in the domain of $G$, namely $input_G$. Therefore:
\begin{align}
    |G| + input_G \geq K^*(G) \label{p:upper1}
\end{align}

From \ref{p:pspace} it follows that:
\begin{align}
      input_G = log_2(\#\pi(G))\label{p:upper2}
\end{align}

From \ref{p:upper1} and \ref{p:upper2}, we can derive $|G| + log_2(\#\pi(G)) \geq K^*(G)$ as required.


\textbf{Lower bound}: $K^*(G) \geq log_2(\#\pi(G))$.

By the definition of Kolmogorov complexity, for each artefact $a$ in $\pi(G)$ there exists at least one deterministic program $p_a$ whose output is $a$, such that $|p_a| = K(a)$. Let $P_a$ be the nonempty set of all such programs. By the definition of $P_a$, we know there are at least as many programs in $P_a$ as there are artefacts in $\pi(G)$, i.e. $|p_a| \geq \#\pi(G)$. As such, we have:
\begin{align}
log_2(|p_a|) \geq log_2(\#\pi(G)) \label{p:lower1}
\end{align}


$P_a$ is a non-empty set of finite binary strings. By the pigeonhole principle, there exists at least one $p \in P_a$ such that: \begin{align}
|p| \geq log_2(|p_a|) \label{p:lower2}
\end{align}

From \ref{p:lower1} and \ref{p:lower2}, and the commutativity of the $\geq$ operator, we derive:
\begin{align}
|p| \geq log_2(\#\pi(G)) \label{p:lower3}
\end{align}

From the definition of $K^*$ in \ref{p:kstar}, we obtain:
\begin{align}
K^*(G) = Max \{|p| : a \in \pi(G), p \in P_a\} \label{p:lower4}
\end{align}

From \ref{p:lower3} and \ref{p:lower4}, it follows that:
\begin{align}
K^*(G) \geq log_2(\#\pi(G))
\end{align}

\end{proof}

\section{Discussion and Implications}

In the previous section we outlined a proof that used Kolmogorov complexity to relate different properties of a generative system to one another, namely the size of its possibility space ($log_2(\#\pi(G))$), the Kolmogorov complexity of its most complex artefact ($K^*(G)$), and the length of the generator's code ($|G|$). In this section we contextualise this result by linking it to aspects of procedural generation practice, and discussing implications of the result for research into PCG.

\subsection{Tradeoffs in Generative Design}
By relating the elements of the proof above to more plain-language, everyday aspects of designing procedural generators, we can begin to link the results of the proof to an intuitive understanding of the limitations of generative systems. 

\subsubsection{Encoded Knowledge}
The design of procedural generators often involves research, practice and experimentation. In order to design a system which procedurally generates poetry, for example, a generative systems designer might read writing about the theory and techniques poets use, try writing poems themselves, or read a lot of poems in a genre or style they are interested in replicating. Such intensive research is not always required -- a designer might decide to base their work on the knowledge of poetry they already have, or might be an experienced poet themselves and already have many years of practice. Even in the case where research or practice is not applicable, for example in using procedural generation as a form of compression or randomisation, care and planning is still required to think about the distribution, function and goals of the generative system. 

In writing a procedural generator, the designer is embedding their knowledge about the generative problem in question into the code they are writing. The more knowledge they wish to embed into the system, the more code they need to write. For example: code to handle edge cases that they wish to exclude from the possibility space; code to describe templates for particular forms or structures; or code to describe particular distributions of noise or randomness to provide the right textural basis. We can think of the length of the generator's code ($|G|$) as an analogue for the design knowledge that has been encoded into the generator. Note that $|G|$ represents \textit{minimal} code, which impacts the generator's functionality, rather than measuring any code at all. Thus, adding empty statements does not increase $|G|$, but adding code which affects how content is generated (thus affecting its Kolmogorov complexity) does count.

\subsubsection{Scale}
Marketing for games which prominently feature procedural generation may also mention the scale or size of the possibility space, such as Borderlands 3's marketing campaign which highlighted that the game contained `over one billion guns'\footnote{https://www.youtube.com/watch?v=bFLhcoFAJMQ}. Many of the early arguments for using procedural generation in game design stemmed from their supposed ability to create `replayability' or `endless' amounts of content for players to consume. While this is certainly true for some uses of the technique, scale is not always needed, nor does it always guarantee quality or fitness. 

In our proof, the size of the possibility space ($log_2(\#\pi(G))$) captures the number of potential outputs the generator can create. A small number of potential outputs might point to a lack of variety in the generator, which might mean the experience of the content suffers from repetition. Alternatively, it might be that the generator is designed to create a specific set of outputs (such as the use of PCG-as-compression in Elite \cite{elite}), or the generator is intentionally kept small so the player can learn or predict its behaviour. Equally, a large number of potential outputs might indicate a bland space of very similar outputs (the \textit{bowls of oatmeal} problem referenced in section \ref{sec:oatmeal}). Alternatively, it might be used to convey vastness and repetition (as suggested by Emily Short \cite{parrigues}), or supported by enough encoded knowledge to maintain diversity even at scale. We discuss the Oatmeal problem, and its relation to our proof, in greater detail in the next section.

\subsubsection{Pattern Density}
Players naturally learn to identify patterns in game content over time. This is not exclusive to procedurally generated content; players often complain about the reuse of assets in multiple areas of a game, for example, or identify the look or feel of a particular game engine. Due to its algorithmic nature, however, procedurally generated content is more susceptible to pattern identification in this way. Generated content might be described as `repetitive' if it is too easy to notice patterns.

There are many approaches to delaying the player's ability to learn patterns. One is to simply add more \textit{pattern density} -- to add more detail and more patterns to the generative processes, so it becomes harder to remember the last time a particular design element was encounter. Another is to add confounding elements that blur the edges between patterns. This approach is used by Spelunky to blend its discrete templates with more continuous randomness. 

Kolmogorov complexity expresses how much program code is required to describe a particular object. The more complex, noisy, detailed or random an artefact is, the more code is required to describe it. The \textit{most} complex artefact in a possibility space (denoted $K^*(G)$ in our proof) represents the ceiling of this property for a particular generator. The higher this value, the more complexity a generator is capable of producing.

\begin{figure}[t]
\begin{minipage}{0.45\textwidth}
\begin{tikzpicture}


\fill[fill=gray!50] plot[smooth, samples=100, domain=0:2] (\x, 2 - \x) -| (0,0) -- cycle;

\fill[fill=gray!50] plot[smooth, samples=100, domain=2:5] (\x, 4) -| (2,0) -- (4,0) -- (4,4);

\draw[->] (-0.2, 0) -- (4.2, 0) node[right] {$log_2(\#\pi(G))$};
\draw[->] (0, -0.2) -- (0, 4.2) node[above] {$|G|$};


\draw[draw, dashed] (2, 0) -- (0, 2) node[left] {$K^*(G)$};
\draw[draw, dashed] (2, 0) node[below] {$K^*(G)$} -- (2, 4);

\draw [fill] (2,2) circle [radius=0.05] node[left] {$Q$};
\draw [fill] (1,1) circle [radius=0.05] node[right] {$P$};

\end{tikzpicture}
\captionof{figure}{A plot of the relationship between $|G|$, the length of the generator, and $log_2(\#\pi(G))$, the size of the possibility space. The intersections marked $K^*(G)$ represent the Kolmogorov Complexity of the most complex artifact in $\pi(G)$.}
\label{fig:kcplot}
\end{minipage}
\hfill
\begin{minipage}{0.45\textwidth}
\begin{tikzpicture}

\draw[->] (-0.2, 0) -- (4.2, 0) node[right] {$log_2(\#\pi(G))$};
\draw[->] (0, -0.2) -- (0, 4.2) node[above] {$|G|$};

\draw[draw, blue, dashed] (3.5, 0) -- (0, 3.5) node[left] {$K^*(F_{12})$};
\draw[draw, blue, dashed] (3.5, 0) node[below] {$K^*(F_{12})$} -- (3.5, 4) ;
\draw [fill=blue,stroke=blue] (2.5,2) circle [radius=0.05] node[above,blue] {$F_{12}$};

\draw[draw, red, dashed] (1, 0) -- (0, 1) node[left] {$K^*(F_6)$};
\draw[draw, red, dashed] (1, 0) node[below] {$K^*(F_6)$} -- (1, 4) ;
\draw [fill=red,stroke=blue] (0.5,2) circle [radius=0.05] node[above,red] {$F_6$};

\draw [-to, thick](2.25,2) -- (0.75,2);
\end{tikzpicture}
\caption{A replot of Figure \ref{fig:kcplot} with marks indicating the two versions of the Flower Generator, $F_{12}$ and $F_6$.}
\label{fig:teagardenplot}
\end{minipage}
\end{figure}

\subsection{Relationship to the Theorem}
Reconsidering the results of the proof in light of these definitions, we can see that there is a relationship between the amount of information encoded within the generative algorithm, the number of things the algorithm can generate, and the complexity of patterns and details within its outputs. Because these concepts are all linked in our theorem, changing any one of them will affect the value of the others.

To visualise this, in Figure \ref{fig:kcplot} we plot the two inequalities in our theorem for a generator $G$. By our theorem, the values of $|G|$ and $log_2(\#\pi(G))$ are bounded by these lines, meaning that $G$ will always be plotted in the unshaded area on the graph. A consequence of this is that attempting to change one of these values may require other values to be changed as well. For example, consider the generator $P$ in the plot. In order to decrease the length of its code ($|G|$) we must either \textit{reduce} $K^*(G)$ or \textit{increase} $log_2(\#\pi(G))$, because of the bounds expressed by our theorem. In the parlance of this section, we cannot remove knowledge from the generator without either reducing the pattern density of what it can generate, or increasing the scale of its output. Similarly for $Q$, increasing $log_2(\#\pi(G))$ requires we increase $K^*(G)$ or, in other terms, in order to increase the scale of the generator's output, we must also increase the pattern density of what it creates. Such invariant relationships between these properties of a procedural generator are explored in greater depth with examples in the following section.

\section{Examples}

\subsection{Flower Generator}
\begin{figure} [t]
\centering

\begin{minipage}{0.45\textwidth}
  \centering
  \includegraphics[width=0.9\linewidth]{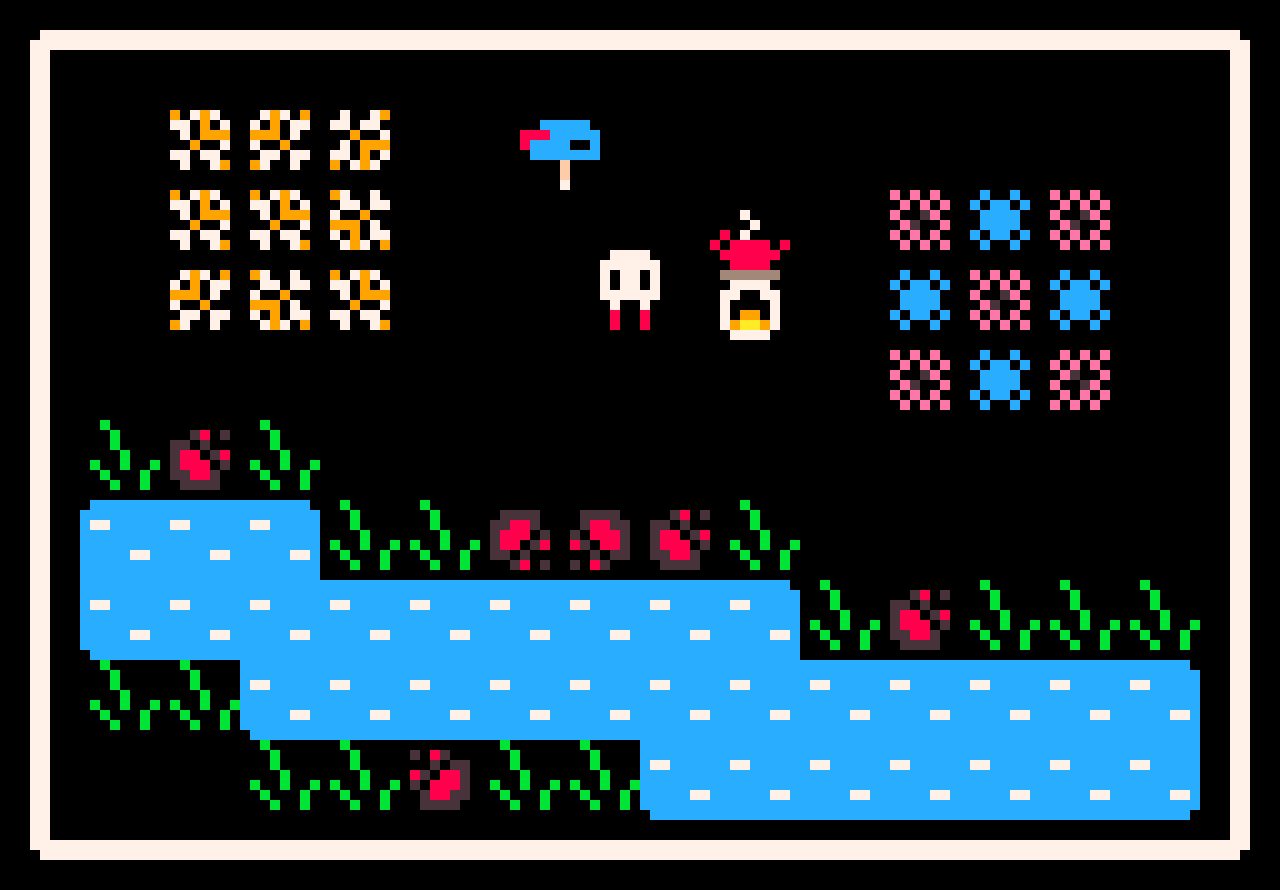}
  \captionof{figure}{A development screenshot of \textit{Tea Garden}. Groups of generated flower sprites can be found at the top right and top left of the map, as well as near the river at the bottom.} 
  \label{fig:teagarden_screen}
\end{minipage}    %
\hfill
\begin{minipage}{0.45\textwidth}
  \centering
  \includegraphics[width=\linewidth]{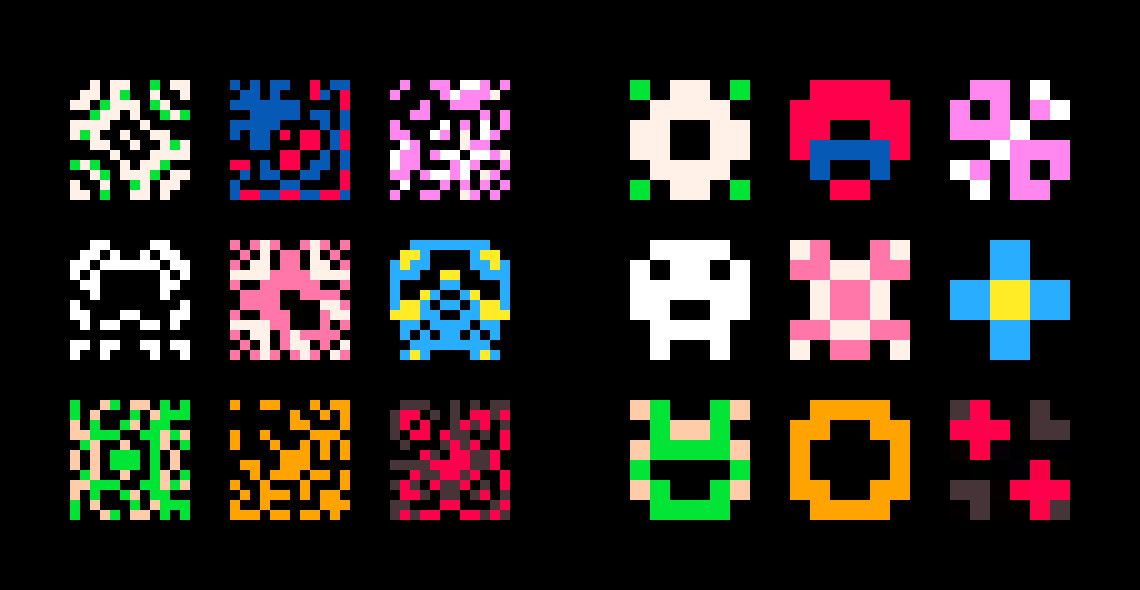}
  \captionof{figure}{Output samples from the 12x12 (left) and 6x6 (right) flower sprite generators, produced during the development of \textit{Tea Garden}. The designer chose to only keep 6x6 artefacts in the final game.}
  \label{fig:teagarden_flowers}
\end{minipage}
\end{figure}

\textit{Tea Garden} is a forthcoming independently-developed videogame about exploring dream worlds. In the game, the player can pick flowers to brew tea, which when drunk induces dreams of gardens full of flowers. The player can pick a single flower to take back out of the dream and into the real world, which can then be used to brew more tea. The game extensively uses procedural generation, in particular to generate both the layout of the dream gardens, and the designs of the flowers themselves. Figure \ref{fig:teagarden_screen} shows a screenshot from the game. 

Consider a procedural content generator that produces flower designs for Tea Garden. The output of the generator is always a two-dimensional array of pixel colours, measuring 12x12. We refer to this generator as $F_{12}$, with the numeric subscript defining the width and height of the flowers in pixels. Now consider a modification of this generator, $F_6$, which is identical save for the size of the flowers, now measuring 6x6. The length of the generator's code has not changed, since we have simply changed one variable describing output size. However, the size of the generator's possibility space has now decreased, because there are fewer flowers that can be represented in 6x6 pixels than there are 12x12. Figure \ref{fig:teagardenplot} shows these two generators on the same plot from Figure \ref{fig:kcplot}. Note how $F_6$ has moved down the x-axis, indicating a smaller possibility space, but has the same spot on the y-axis, as the length of the code has not changed.

We can see from this that a consequence of this change is that the most complex artifact the generator can produce is also reduced. The overall effect of this is that we have reduced the size and complexity of the generator's output, without reducing the amount of knowledge encoded in it. This focuses the generator around a smaller set of outputs. Note that this does not necessarily result in a `better' generator, but it does result in a generator whose output is slightly easier to understand and describe, because it contains less information and variation within it. In this case, \textit{Tea Garden}'s designer preferred to replace $F_{12}$ with $F_6$ for their final design, as the reduced complexity of the generated sprites better suited the game's low resolution pixelart aesthetic. Figure \ref{fig:teagarden_flowers} shows a comparison between $F_{12}$ and $F_6$'s outputs.

\subsection{Minecraft}
\textit{Minecraft} is a 3D survival crafting game, and one of the most popular games to prominently feature procedural generation. \textit{Minecraft} is set inside procedurally generated worlds that are far larger than any player could ever explore, using an algorithm that has been carefully iterated upon over many years to create dramatic, interesting and beautiful worlds that also serve important gameplay functions such as providing challenge, inspiration and surprise. The history of the \textit{Minecraft} world generator can be seen from its patch notes and community records \cite{minecraft_wiki}. We refer to each version of the generator here as $M_n$ where $n$ is the major version number of Minecraft associated with it. 

Between $M_{1.7}$ and $M_{1.8}$, \textit{Minecraft} designers added villages to its world generator. During the generation process, the generator will mark an area to have a village placed in it, and then use templates for houses, farms and other structures to construct a village. Figure \ref{fig:minecraftplot} shows $M_{1.7}$ and $M_{1.8}$ on the same plot as Figure \ref{fig:kcplot}. 

Since most of the village's blocks are put in space that would have been empty otherwise, we consider this feature is evidence that \textit{Minecraft}'s designers desired to augment the complexity of its generated worlds ($K^*(G)$). As both versions of Minecraft used a 64-bit seed leading to a maximum of $2^{64}$ generated worlds, the two generators have the same scale ($log_2(\#\pi(G))$), putting them on the same spot on the x-axis. The theorem predicts that if the increase of complexity between $M_{1.7}$ and $M_{1.8}$ is high enough, the length of \textit{Minecraft}'s source code will have to increase too.

In that likely case, \textit{Minecraft}'s designers would have increased the complexity of their worlds in exchange for encoding more designer knowledge in its world generator. The requirement to invest skills and resources in a generator to raise the complexity of its artefacts is supported by the existence of PCG design competitions dedicated to generating villages in Minecraft, like the GDMC AI Settlement Generation Challenge.

\begin{figure}[t]
\begin{minipage}{.45\textwidth}
\begin{tikzpicture}
\draw[->] (-0.2, 0) -- (4.2, 0) node[right] {$log_2(\#\pi(G))$};
\draw[->] (0, -0.2) -- (0, 4.2) node[above] {$|G|$};

\draw[draw, blue, dashed] (1.5, 0) -- (0, 1.5) node[left] {$K^*(M_{1.7})$};
\draw[draw, blue, dashed] (1.5, 0) node[below] {$K^*(M_{1.7})$} -- (1.5, 4) ;
\draw [fill=blue,stroke=blue] (1,1.5) circle [radius=0.05] node[above,blue] {$M_{1.7}$};

\draw[draw, red, dashed] (3.5, 0) -- (0, 3.5) node[left] {$K^*(M_{1.8})$};
\draw[draw, red, dashed] (3.5, 0) node[below] {$K^*(M_{1.8})$} -- (3.5, 4) ;
\draw [fill=red,stroke=blue] (1,3.5) circle [radius=0.05] node[above,red] {$M_{1.8}$};

\draw [-to, thick](1,2) -- (1,3.25);

\end{tikzpicture}
\captionof{figure}{A plot showing the change between two versions of the Minecraft world generator, $M_{1.7}$ and $M_{1.8}$.}
\label{fig:minecraftplot}
\end{minipage}
\hfill
\begin{minipage}{.45\textwidth}
\begin{tikzpicture}
\draw[->] (-0.2, 0) -- (4.2, 0) node[right] {$log_2(\#\pi(G))$};
\draw[->] (0, -0.2) -- (0, 4.2) node[above] {$|G|$};

\draw[draw, blue, dashed] (1.5, 0) -- (0, 1.5) node[left] {$K^*(G_{1})$};
\draw[draw, blue, dashed] (1.5, 0) node[below] {$K^*(G_{1})$} -- (1.5, 4) ;
\draw [fill=blue,stroke=blue] (1,1.5) circle [radius=0.05] node[above,blue] {$G_{1}$};

\draw[draw, red, dashed] (3.5, 0) -- (0, 3.5) node[left] {$K^*(G_{2})$};
\draw[draw, red, dashed] (3.5, 0) node[below] {$K^*(G_{2})$} -- (3.5, 4) ;
\draw [fill=red,stroke=blue] (2.5,1.5) circle [radius=0.05] node[above,red] {$G_{2}$};

\draw [-to, thick](1.25,1.5) -- (2.25,1.5);
\end{tikzpicture}

\captionof{figure}{A plot showing the change in a hypothetical generator, leading to the generation of `oatmeal'.}
\label{fig:oatmealplot}

\end{minipage}

\end{figure}

\subsection{The Cost of Oatmeal}

In section \ref{sec:oatmeal} we described the \textit{ten thousand bowls of oatmeal} problem, where a large quantity of content is produced, but the quality and variety of the content is so low that the quantity becomes a problem rather than a boon. In this example we explore how this can happen unintentionally when attempting to engineer a more complex generator.

Suppose we have developed a generator, $G_1$ in Figure \ref{fig:oatmealplot}, and we wish to make its output more complex -- that is, we wish to increase the pattern density, $K^*(G_1)$. Increasing $K^*$ will eventually require us to change either the length of the program ($|G|$) or the size of the possibility space ($log_2(\#\pi(G))$), since our theorem guarantees that the inequality $|G| + log_2(\#\pi(G)) \geq K^*(G)$ holds for any ideal generator $G$. As a developer, this offers us two solutions. The first is to add to the length of the program, adding more encoded knowledge into the generator. However, this solution is both costly and time-consuming. The second solution is to increase the size of the possibility space and scale up the generator, for example by randomly combining subcomponents of our artefacts.  This is a cheap, and therefore appealing, solution, and results in a generator such as $G_2$ in Figure \ref{fig:oatmealplot}. By linearly increasing $K^*$ in this way however, we are \textit{exponentially} increasing the size of the possibility space without adding any new encoded knowledge to control or shape output. This is highly likely to result in a large quantity of noisy, perceptually similar, unremarkable content, otherwise known as \textit{oatmeal}. We claim that a common reason for the oatmeal phenomenon is that \textit{oatmeal is cheap}, as it does not require the costly encoding of knowledge in order to increase the pattern density of a generator.


\section{Future Work}
This paper presents a first step in relating ideas from complexity theory to generative systems. There are many ways in which this work can be extended to increase the topics it covers, or to explore new applications. For example, Kolmogorov Complexity was not designed with neural networks in mind, but the burgeoning field of Procedural Content Generation via Machine Learning makes this an important area of the field to consider \cite{pcgml}. We aim to investigate how Kolmogorov Complexity can be used to express constraints on PCGML systems in the future, too.

\section{Conclusions}
In this paper we introduced the concept of Kolmogorov Complexity from information theory and related it to the study of procedural content generation. Specifically, we provided a proof of the relationship between the length of a generator's source code, the size of its possibility space, and the highest Komogorov Complexity the generator is capable of producing. We then argued that this conveys a well-understood tradeoff in procedural generation practice, between the scale of a procedural generator's outputs, how dense or noisy the space is, and how detailed its generative algorithm is. We then used several real-world examples to show how this idea can be applied to generative systems.

\begin{acks}
We would like to thank Azalea Raad for her feedback on our proof, as well as our reviewers for taking the time to provide us with constructive feedback.
\end{acks}

\appendix
\section{Transforming Non-Ideal Generators} \label{sec:transform}
Let $G$ be a non-ideal generator which is terminable, but does not does not have a fixed input size (but admits a finite set of inputs and does not admit infinite inputs) and is not injective. Let $I$ be the domain of $G$ (i.e.~ its set of inputs) and let $m$ be the length of the longest $i \in I$.

Let us define $enc_m$ as a function that takes as input a binary string, $i \in I$, and returns $|i|$ expressed as a binary string, denoted by $b_i$, padded with leading zeroes if $|b_i|$ is less than $\lceil{log_2(m)}\rceil$, concatenated with $i$, padded with leading zeroes if $|i|$ is less than $m$. 
Let $I'' = \{enc_m(i) : i \in I\}$;
note that $enc_m(i)$ is an \emph{injective} function in that it transforms each $i \in I$ into a unique string $i'' \in I''$ of length $m  +\lceil{log_2(m)}\rceil$.

Let $G''$ be a generator with domain $I''$ such that $G''(i) = G(enc_m^{-1}(i))$, for all $i \in I''$ (as $enc_m$ is injective, $enc_m^{-1}$ is well-defined).
The generator $G''$ then satisfies the fixed input size property as every string in $I''$ is of fixed length $m  +\lceil{log_2(m)}\rceil$ and the domain of $G''$ is $I''$.

Let $G'$ be a generator with domain $I''$ such that $G'(i) = G''(i) + i$, for all $i \in I''$. 
That is, the output of $G'$ given input $i$ is the output of $G''$ given input $i$, concatenated with $i$ itself. Note that $G'$ is an injective function: for any two inputs $i, j \in I''$, if $i \ne j$, then $G'(i) = G''(i) + i$ and $G'(j) = G''(j) + j$, and thus $G'(i) \ne G'(j)$.
Moreover, as the domain of $G'$ is $I''$, it also satisfies the fixed input size. 
As such, $G'$ is an ideal generator.

\bibliographystyle{ACM-Reference-Format}
\bibliography{sample-base}

\end{document}